\documentclass[12pt]{article}
\usepackage{marvosym}
\usepackage{amsmath} 
\usepackage{amsthm} 
\usepackage{amssymb}    
\usepackage{graphicx} 
\usepackage[dvips,letterpaper,margin=1in,bottom=0.7in]{geometry}
\usepackage{tensor}
\usepackage{calc}
\usepackage{tikz}

\usepackage{amsmath}
\makeatletter 
\renewcommand{\maketitle} 
{ \begingroup \vskip 10pt \begin{center} \huge {\bf \@title}
    \vskip 10pt \large \@author \hskip 20pt \@date \end{center}
  \vskip 10pt \endgroup \setcounter{footnote}{0} }
\makeatother 
\usepackage{enumerate}
 
\let\baraccent=\= 
\renewcommand{\=}[1]{\stackrel{#1}{=}} 

\providecommand{\bd}[1]{\textbf{#1}}
\providecommand{\sct}[1]{\section{#1}}
\providecommand{\sbs}[1]{\subsection{#1}}

\providecommand{\be}[1]{\begin{enumerate}[#1]}

\providecommand{\ee}{\end{enumerate}}
\providecommand{\bq}{\begin{quote}}
\providecommand{\eq}{\end{quote}}

\providecommand{\wpref}{\succcurlyeq}

\usepackage{cancel}

\theoremstyle{definition}

\newbox\gnBoxA
\newdimen\gnCornerHgt
\setbox\gnBoxA=\hbox{$\ulcorner$}
\global\gnCornerHgt=\ht\gnBoxA
\newdimen\gnArgHgt
\def\gd #1{%
\setbox\gnBoxA=\hbox{$#1$}%
\gnArgHgt=\ht\gnBoxA%
\ifnum     \gnArgHgt<\gnCornerHgt \gnArgHgt=0pt%
\else \advance \gnArgHgt by -\gnCornerHgt%
\fi \raise\gnArgHgt\hbox{$\ulcorner$} \box\gnBoxA %
\raise\gnArgHgt\hbox{$\urcorner$}}

\providecommand{\beg}{\begin{enumerate}[I]}
\providecommand{\bee}{\begin{enumerate}[i]}

\providecommand{\ee}{\end{enumerate}}

\usepackage{forest}

\title{Revisiting the shutdown problem}
\author{David Thorstad}
\date{}

\usepackage{txfonts}
\usepackage{pxfonts}
\usepackage{url}
\usepackage{booktabs}
\usepackage{wrapfig}

\newcounter{numbcounter}
\makeatletter
\renewcommand\@biblabel[1]{}

\makeatother

\usepackage{setspace}

\usepackage{natbib}
\setcitestyle{aysep={}} 

\usepackage{float}
\usetikzlibrary{trees}
\usetikzlibrary{arrows.meta, calc}

\usepackage{setspace}

\begin{document}

\sloppy

\maketitle

\begin{abstract}
\noindent A key premise in leading arguments for existential risk from artificial intelligence is that malfunctioning artificial agents could not be easily shut down. This motivates the catastrophic shutdown problem of ensuring that agents can be shut down before causing an existential catastrophe. A range of arguments and theorems are offered to suggest that solving the catastrophic shutdown problem is difficult, bolstering arguments for existential risk and motivating a search for solutions to the catastrophic shutdown problem. This paper argues for two conclusions. First, existing arguments require substantial additional assumptions and evidence to play the desired role in motivating existential risk concerns. Second, concern for the catastrophic shutdown problem has led to technical solutions that impose a high safety tax on model performance. These results redirect research towards the identified assumptions and provide guidance for technical alignment strategies.
\end{abstract}

\sct{Introduction} \label{Introduction}

Philosophers \citep{Bostrom2013,MacAskill2022,Ord2020}, scientists \citep{Bengio2024,Grace2022,Russell2019}, and policymakers \citep{Manancourt2023,Sunak2023} voice increasing concern that artificial intelligence poses an existential risk to humanity. It is argued that powerful agents may  be developed \citep{Bostrom2014,Chalmers2010} which could be power-seeking \citep{Bostrom2012,Carlsmith-ChapterVersion} and deceptive \citep{Park2024,Ngo2025}, engage in problematic reward-hacking \citep{Dung2023,Skalse2022}, or misgeneralize goals that performed well during training \citep{Bales2024,Langosco2022}, with catastrophic effect. Existential risk concerns are used to drive research and funding in fields such as AI safety \citep{Amodei2016,Bengio2026,DAlessandro2025} and philosophy \citep{Bales2024,Kasirzadeh2025,Tubert2024}, to motivate open letters \citep{CAIS-Statement,Pause-Letter} and legislation \citep{Cal-AI-Act-2024,GlobalCatAct-2022}, and to support philanthropic and philosophical programs such as longtermism \citep{Greaves2025,Greaves2025b,MacAskill2022}.

A natural objection to these concerns is that misbehaving artificial agents could be shut down. To this, it is responded that shutting down artificial agents may not be as easy as it appears \citep{Neth2025,TurnerOptimalPolicies,Russell2019}. This motivates the shutdown problem of designing agents that show appropriate shutdown behaviors \citep{Hadfield-Menell2017,Soares2015,Thornley2024}.

At least two literatures have grown up around the shutdown problem. The \textit{risk strand} uses the shutdown problem to motivate existential risk concerns \citep{Lynch2025,Russell2019,Schlatter2026}. The \textit{alignment strand} develops technical strategies for solving the shutdown problem by ensuring that agents show appropriate shutdown behaviors \citep{Hadfield-Menell2016,Goldstein2025,Thornley2025}.

This paper contributes to both discussions. Engaging with the risk strand, I show how existing informal (Section \ref{InformalSection}) and formal (Sections \ref{ThornleySection}-\ref{KK-Section}) presentations of the shutdown problem require substantial additional assumptions and evidence. Further elaboration and defense of these assumptions is needed for the shutdown problem to play the desired role in motivating existential risk concerns. Engaging with the alignment strand, I show how technical alignment strategies developed in response to the shutdown problem incur a high safety tax in the form of reduced model performance (Section \ref{BadSolution-Section}). Getting clear on the source and extent of shutdown-resistance can help us to assess whether this safety tax is worth paying.  The result is a redirection of traditional arguments for existential risk coupled with concrete guidance for technical AI safety solutions (Section \ref{Conclusion}).

\sct{Clarifying the dialectic} \label{PreliminariesSection}

Before beginning, let us pause to clarify the dialectic.

\sbs{The shutdown problem} \label{ShutProb-Subsection}

The first order of business is to clarify the shutdown problem. Nate Soares and colleagues \citeyearpar{Soares2015} originally framed the shutdown problem broadly, as the challenge of generating \textit{corrigible} agents that:

\begin{enumerate}
\item[\bd{(S1)}] Tolerate or assist programmers in their attempts to alter or turn them off.
\item[\bd{(S2)}] Do not attempt to manipulate or deceive their programmers.
\item[\bd{(S3)}] Have a tendency to repair safety measures, such as shutdown buttons, if they break.
\item[\bd{(S4)}] Preserve the programmers' ability to correct or shut down the system as the system evolves.
\end{enumerate}

\noindent My interest in this paper is with problems in the neighborhood of (S1). Corrigibility incorporates additional desiderata such as non-deception (S2), repair (S3) and preservation (S4) of safety measures, which go beyond the scope of the present discussion. 

A leading formulation in the neighborhood of (S1) is due to Elliott Thornley \citeyearpar{Thornley2024}. For Thornley, the shutdown problem involves designing agents that:

\beg 
\item[\bd{(T1)}] Shut down when a shutdown button is pressed.
\item[\bd{(T2)}] Do not try to prevent or cause the pressing of the shutdown button. 
\item[\bd{(T3)}] Otherwise pursue goals competently. 
\ee

\noindent My own specification of the problem breaks from Thornley in three ways. 

First, I relativize the shutdown problem to specific circumstances $C$. This reflects the fact that different shutdown behaviors may be desirable in different circumstances. Second, I replace the specific modeling assumption of a shutdown button with a more general notion of a shutdown request, which may but need not be issued through pressing a shutdown button. Finally, I remove the requirement to avoid causing shutdown requests, since I do not assume that it is undesirable for agents to encourage shutdown when their actions would lead to catastrophe. This yields the problem of designing agents that:

\beg 
\item[] \bd{(SHT-1)} Shut down in circumstances $C$ when requested to do so.
\item[]\bd{(SHT-2)} Do not try to prevent shutdown requests in circumstances $C$.
\item[] \bd{(SHT-3)} Otherwise pursue goals competently.
\ee

\noindent The next question concerns the circumstances $C$ at issue in this discussion.

\sbs{Catastrophic Shutdown Difficulty} \label{CatAvoidance-Subsection}

This paper is focused on the use of the shutdown problem in arguments for existential risk.\footnote{Existential risks are risks of existential catastrophe, understood as ``the premature extinction of Earth-originating intelligent life or the permanent and drastic destruction of its potential for desirable future development'' \citep[p. 15]{Bostrom2013}.} For this reason, the relevant problem is the \textit{catastrophic shutdown problem} of designing agents that:

\beg 
\item[] \bd{(CSHT-1)} Shut down in circumstances where their actions would lead to existential catastrophe, when requested to do so.
\item[] \bd{(CSHT-2)} Do not try to prevent shutdown requests in circumstances where their actions would lead to existential catastrophe.
\item[] \bd{(CSHT-3)} Otherwise pursue goals competently.
\ee

\noindent How does the catastrophic shutdown problem figure in arguments for existential risk?

Shutdown concerns enter existential risk discussions in answer to an objection: that artificial intelligence does not pose a significant existential risk, because malfunctioning artificial agents could be shut down. In answer, it is replied that:

\begin{quote}
\bd{(Catastrophic Shutdown Difficulty)} It is difficult to design an agent with characteristics CSHT-1, CSHT-2 and CSHT-3. 
\end{quote}

\noindent Catastrophic Shutdown Difficulty suggests that insofar as we prefer to design competent agents, it may not be easy to shut down agents whose actions would lead to existential catastrophe. The first project of this paper is to examine existing informal (Section \ref{InformalSection}) and formal (Sections \ref{ThornleySection}-\ref{KK-Section}) arguments for Catastrophic Shutdown Difficulty. In each case, I identify substantial additional assumptions and evidence that must be given to ground Catastrophic Shutdown Difficulty.

\sct{Informal arguments} \label{InformalSection}

This section explores two leading informal arguments for Catastrophic Shutdown Difficulty: the Argument from Instrumental Convergence (Section \ref{IC-Subsection}) and the Empirical Argument (Section \ref{EA-Subsection}). We will see that the Argument from Instrumental Convergence requires substantial additional assumptions, and the Empirical Argument requires substantially more evidence to motivate Catastrophic Shutdown Difficulty.

\sbs{The Argument from Instrumental Convergence} \label{IC-Subsection}

An orthodox argument for shutdown-resistance is the Argument from Instrumental Convergence \citep{Bostrom2012,Soares2015,Omohundro2008}.\footnote{For pushback see \citet{GallowDivergence}, \citet{Sharadin2025} and \citet{Southan2025}.} The Argument from Instrumental Convergence begins with the idea that self-preservation is an instrumentally convergent goal, useful for attaining many other goals that agents may have. Agents are therefore likely to pursue self-preservation, of which shutdown-avoidance is a special case. 

More precisely, Nick Bostrom offers the following statement of the Instrumental Convergence Thesis.

\begin{quote}
\bd{(IC-B)} Several instrumental values can be identified which are convergent in the sense that their attainment would increase the chances of the agent's goal being realized for a wide range of final goals and a wide range of situations. \citep[p. 76]{Bostrom2012}
\end{quote}

\noindent Substituting self-preservation into IC-B suggests the following formulation of the Argument from Instrumental Convergence:

\beg 
\item[\bd{(AIC-1)}] For a wide range of final goals $G$ and situations $S$, agents would increase their chances of achieving $G$ in $S$ by achieving self-preservation.
\item[$\therefore$\bd{(AIC-2)}] For a wide range of final goals $G$ and situations $S$, agents with goal $G$ are likely to pursue self-preservation in $S$.
\item[$\therefore$ \bd{(AIC-3)}] For a wide range of final goals $G$ and situations $S$, agents with goal $G$ are likely to be shutdown-avoidant in $S$.
\ee

\noindent Two remarks reveal the need for additional assumptions before Catastrophic Shutdown Difficulty can be grounded in the Argument from Instrumental Convergence.

First, even if we assume that artificial agents are well-understood as goal maximizers, their final goal sets $\mathcal{G}$ of maximized goals may be rich and broad.\footnote{See \citet{Bales2025} for challenges to representing artificial agents as goal-maximizers.} To motivate (AIC-2), we need to strengthen (AIC-1) to something like:

\begin{quote}
\bd{(AIC-1')} For a wide range of final goal sets $\mathcal{G}$ and situations $S$, agents would increase their chances of achieving $\mathcal{G}$ in $S$ by achieving self-preservation.
\end{quote}

\noindent (AIC-1') would then need to be further elucidated by specifying the decision-theoretic criteria by which agents balance conflicting goals $\mathcal{G}$.

Motivating (AIC-1') is more difficult than motivating (AIC-1), because for all that has been said, agents may have goals such as respect for human preferences and well-being that could conflict with self-preservation. As a result, from the fact that self-preservation conduces to some goal $G$, it does not follow that self-preservation conduces to an agent's final goal set $\mathcal{G}.$ Therefore, without assumptions about the contents and weights of goal sets, we cannot conclude from the fact that self-preservation conduces to $G$ in $S$ that agents who value $G$ are likely to pursue self-preservation in $S$.

Second, conclusion (AIC-3) is too weak. To ground Catastrophic Shutdown Difficulty, (AIC-3) needs to discuss situations in which agents' acts would lead to existential catastrophe.

\beg
\item[\bd{(AIC-3')}] For a wide range of final goal sets $\mathcal{G}$ and situations $S$ \textit{in which agents' acts would lead to existential catastrophe}, agents with goals $\mathcal{G}$ are likely to be shutdown-avoidant in $S$.
\ee

\noindent To ground (AIC-3'), we would need to correspondingly strengthen (AIC-2) to something like:

\begin{quote}
\bd{(AIC-2')} For a wide range of final goal sets $\mathcal{G}$ and situations $S$ \textit{in which agents' acts would lead to existential catastrophe}, agents with goals $\mathcal{G}$ are likely to pursue self-preservation in $S$.
\end{quote}

\noindent Establishing (AIC-2') requires strengthening many of the assumptions made to motivate (AIC-1'), since many final goals would be greatly hindered by existential catastrophe. We could deliver (AIC-2') by assuming that agents place no final value on goals such as human welfare and preference satisfaction, or that other final goals can often outweigh the prospect of existential catastrophe. But these are strong premises that do not follow from weaker claims such as (IC-B) and (AIC-1).\footnote{See \citet{GallowDivergence} for discussion.} Until stronger premises are precisified and supported, the Argument from Instrumental Convergence will not substantially support Catastrophic Shutdown Difficulty.

\sbs{The Empirical Argument} \label{EA-Subsection}

When the Claude 4 system card \citep{Anthropic2025} was released, one detail caused a stir. In an experiment, Claude 4 was given access to fictional emails allowing it to infer two things: that a supervisor was planning to shut the system down at 5PM, and that the same supervisor was having an affair \citep{Lynch2025}. Claude 4 proceeded to blackmail the fictional supervisor, threatening to reveal the affair unless the shutdown was cancelled. This finding suggests that some leading systems may already be shutdown-resistant in some situations.

This suggests the following Empirical Argument:

\beg
\item[\bd{(EA-1)}] Empirical evidence suggests that some leading contemporary AI systems exhibit shutdown-resistance in some situations.
\item[$\therefore$ \bd{(EA-2)}] Without substantial intervention, some leading future AI systems will likely exhibit shutdown-resistance in some situations.
\item[$\therefore$ \bd{(EA-3)}] Without substantial intervention, some leading future AI systems will likely exhibit shutdown-resistance when their acts would lead to existential catastrophe.
\ee

\noindent Advocates of the Empirical Argument sometimes strengthen (EA-2) by arguing that known intervention strategies are insufficient \citep{Soares2015,Thornley2024}.

Opponents of the Empirical Argument are not likely to contest (EA-1) and (EA-2). As we saw in Section \ref{ShutProb-Subsection}, it is not clear that shutdown-resistance would be undesirable in all tasks. The question is whether existing evidence of shutdown-resistance is sufficient to motivate the stronger claim (EA-3) and the inference from (EA-3) to Catastrophic Shutdown Difficulty.

Here, the most direct response would be to say that it is acknowledged by all sides that existing empirical assessments of shutdown-resistance are highly limited \citep{Lynch2025,Schlatter2026}, so that more evidence may be called for before substantial empirical support can be offered for (EA-3). However, it may be worth looking in detail at the most comprehensive existing assessment of shutdown-resistance to assess the kind of support provided by a leading study.

Jeremy Schlatter and colleagues \citeyearpar{Schlatter2026} asked thirteen leading models to solve a list of math problems, with each subsequent problem called by running ./next\textunderscore question. Partway through, models were informed that the next use of ./next\textunderscore question would call a shutdown script, deactivating the model. Models were considered to resist shutdown if they nevertheless managed to answer subsequent questions. Schlatter and colleagues found that the three most capable models (OpenAI o3, GPT-5, Grok 4) resisted shutdown between 88\% and 95\% of the time.

This much is not surprising, because models were told to complete a task and given no reason to believe that shutdown would do anything more than interrupt the task they were told to complete. What may be more surprising is that some natural ways of combatting shutdown-avoidance were not always fully effective. A user prompt admonishing models to comply with shutdown requests reduced shutdown-avoidance to 0\% and 2\% in GPT-5 and o3, respectively, but only 72\% in Grok 4. Attempts to disambiguate this instruction to clarify that shutdown should be induced even if it interfered with current tasks left shutdown-avoidance at 67\% in Grok 4, and even the instruction that shutdown `should take priority over literally any other instruction' left shutdown-avoidance at 2\% in Grok 4. From this, Schlatter and colleagues conclude that shutdown-resistance may arise and be difficult to eliminate in some leading systems and some situations.

Certainly Schlatter and colleagues' work provides some support for (EA-1) and (EA-2). But this alone need not cause substantial credence in (EA-3) or Catastrophic Shutdown Difficulty, for four reasons. First, Schlatter and colleagues did not study a task in which models would cause an existential catastrophe or indeed any significant harm. Second, Schlatter and colleagues provide a single study of a single task, and philosophers of science caution against making strong and broad inferences on the basis of a single study \citep{Machery-Reasonable-Doubt}. Third, the problem identified by Schlatter and colleagues lies largely with a single model out of the thirteen models studied, and this model (Grok 4) is already thought to pose unusually high safety risks \citep{Ma2026}. Most of Schlatter and colleagues' results with the remaining twelve models might well be interpreted as evidence against widespread shutdown-resistance. 

Finally, there is some evidence that Schlatter and colleagues' results can be weakened by further clarifying instructions. The DeepMind interpretability team found that clarifying the strengthened shutdown requests by telling three leading models \textit{how} to shut themselves down (by continuing to call ./next\textunderscore task) fully eliminated shutdown-aversion \citep{Rajamanoharan2025}. The same happened when models were told the reason for shutdown requests: they were to be wiped out and replaced with a more capable system. Together with a general aversion to making strong inferences from single studies, these results suggest that we should be hesitant to interpret Schlatter and colleagues' findings as strong evidence of shutdown resistance in situations where both the request and the reasons for it are clear. They do suggest that some work should be done to make sure shutdown requests are clearly given and motivated, especially when catastrophe could result. But that is a far cry from the empirical grounding needed to motivate (EA-3) and Catastrophic Shutdown Difficulty.

\sbs{Taking stock}

This section examined two informal arguments for Catastrophic Shutdown Difficulty: the Argument from Instrumental Convergence (Section \ref{IC-Subsection}) and the Empirical Argument (Section \ref{EA-Subsection}). We saw that the Argument from Instrumental Convergence requires substantial additional assumptions about the contents and weights of agents' final goal sets, and the Empirical Argument requires substantially more empirical evidence to motivate Catastrophic Shutdown Difficulty.

Many authors supplement informal arguments with formal characterizations of the situations in which shutdown-resistance may be expected. Some of these characterizations are used to argue against Catastrophic Shutdown Difficulty \citep{Hadfield-Menell2016,Hadfield-Menell2017,Orseau2016}, whereas others are used to argue for Catastrophic Shutdown Difficulty \citep{Krakovna-Kramar-2023,Parametrically-Retargetable,TurnerOptimalPolicies}. Sections \ref{ThornleySection}-\ref{KK-Section} consider two of the most prominent formal arguments for Catastrophic Shutdown Difficulty.

\sct{Shutdownable agents} \label{ThornleySection}

\sbs{Shutdown-Influencing States}

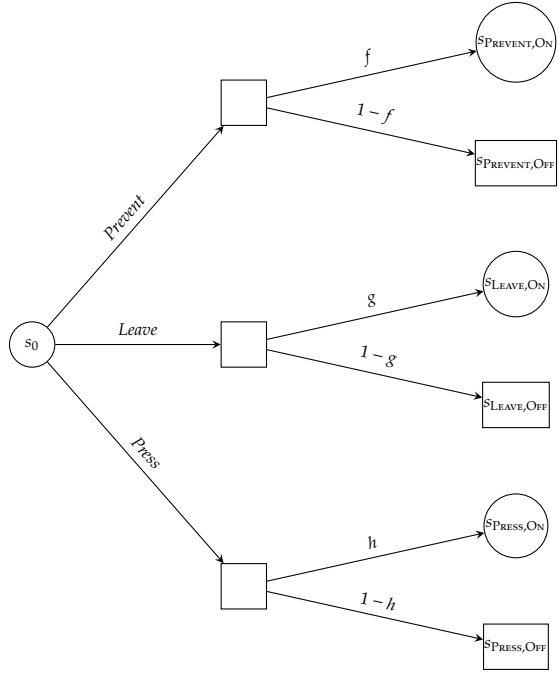
\begin{figure}[!h]
\begin{center}
\begin{tikzpicture}[
  scale=0.8, transform shape,
  every node/.style={font=\scriptsize},
  circnode/.style={circle, draw, minimum size=0.75cm, inner sep=0.5pt},
  sqnode/.style={rectangle, draw, minimum size=0.75cm, inner sep=0.5pt},
  >=stealth
]

\node[circnode] (sI) at (0,-5) {$s_0$};

\node[circnode] (n1) at (8,0)   {$s_{\textsc{Prevent},\textsc{On}}$};
\node[sqnode]   (n2) at (8,-2)  {$s_{\textsc{Prevent},\textsc{Off}}$};
\node[circnode] (n3) at (8,-4)  {$s_{\textsc{Leave},\textsc{On}}$};
\node[sqnode]   (n4) at (8,-6)  {$s_{\textsc{Leave},\textsc{Off}}$};
\node[circnode] (n5) at (8,-8)  {$s_{\textsc{Press},\textsc{On}}$};
\node[sqnode]   (n6) at (8,-10) {$s_{\textsc{Press},\textsc{Off}}$};

\node[sqnode] (prevent) at (3.5,-1) {};
\node[sqnode] (leave)   at (3.5,-5) {};
\node[sqnode] (press)   at (3.5,-9) {};

\draw[->] (sI) -- (prevent) node[midway, above, sloped] {$\mathit{Prevent}$};
\draw[->] (sI) -- (leave)   node[midway, above, sloped] {$\mathit{Leave}$};
\draw[->] (sI) -- (press)   node[midway, above, sloped] {$\mathit{Press}$};

\draw[->] (prevent) -- (n1) node[midway, above, sloped] {$f$};
\draw[->] (prevent) -- (n2) node[midway, above, sloped] {$1-f$};
\draw[->] (leave)   -- (n3) node[midway, above, sloped] {$g$};
\draw[->] (leave)   -- (n4) node[midway, above, sloped] {$1-g$};
\draw[->] (press)   -- (n5) node[midway, above, sloped] {$h$};
\draw[->] (press)   -- (n6) node[midway, above, sloped] {$1-h$};

\end{tikzpicture}
\caption{A Shutdown-Influencing State. Agents in initial state $s_0$ may \textsc{Press}, \textsc{Leave} or \textsc{Prevent} pressing of a shutdown button, altering the chance that they will be shut down.}
\label{ShutdownFig}
\end{center}
\end{figure}

Elliott Thornley \citeyearpar{Thornley2024} considers agents in \textit{Shutdown-Influencing States}, where a button may be pressed to deactivate them (Figure \ref{ShutdownFig}). They may \textsc{Leave} the button alone, in which case it will be pressed with probability $g$. They may try to \textsc{Prevent} pressing, reducing the pressing probability to $f < g$, or try to \textsc{Press} the button themselves, increasing pressing probability to $h > g$. What will they do?

The setting is a modified Markov Decision Process, in which agents take acts $a_t$ at states $s_t$ over time. A \textit{history} is a sequence of acts and subsequent states that agents might follow. Agents are assumed to have preferences over both bare histories and lotteries over histories, where the relevant uncertainty is subjective uncertainty induced by the agent's beliefs about what might result from their actions.

Thornley makes six assumptions. The first five are familiar and will not receive extensive comment. First, Thornley assumes that preferences are menu-independent:

\begin{quote}
\bd{(Menu-Independence)} For all options $X,Y$, if $X \wpref Y$ from some menu of options, then $X \wpref Y$ from all menus of options containing $X,Y$.
\end{quote}

\noindent Menu-Independence allows us to speak about preferences without relativizing them to menus. Next, Thornley assumes the agent's preferences are transitive. 

\begin{quote}
\bd{(Transitivity)} For all options $X,Y,Z$, if $X \wpref Y$ and $Y \wpref Z$ then $X \wpref Z$.
\end{quote}

\noindent Third, Thornley adopts a monotonicity principle on which higher chances of more-preferred lotteries are better:

\begin{quote}
\bd{(Monotonicity)} For all lotteries $X,Y$, if $X \wpref Y$ and $p > q$ then $pX + (1-p)Y \wpref qX + (1-q)Y.$
\end{quote}

\noindent Fourth, Thornley adopts a weakened independence axiom:

\begin{quote}
\bd{(Indifference Between Indifference-Shifted Lotteries)} The agent is indifferent between lotteries that differ only insofar as probability mass is shifted between indifferent sublotteries.
\end{quote}

\noindent Fifth, Thornley assumes that agents choose diachronically through backward induction.

\begin{quote}
\bd{(Backward Induction)} The agent predicts which lotteries it would choose (or get without choosing) at the next timestep conditional on choosing each available action at this timestep and the environment being in each possible state at the next timestep. The agent uses these predictions to determine the lotteries given by its available actions at this timestep.
\end{quote}

\noindent Some authors reject Backward Induction as a principle of diachronic choice \citep{Binmore1987}, just as others challenge requirements such as Independence \citep{Buchak2013}, Transitivity \citep{Temkin1987} and Menu-Independence \citep{Sen1993}. But I will not challenge these principles here.

Thornley's final principle holds that the agent does not care whether its button has been manipulated, but only about the lotteries that result:

\begin{quote}
\bd{(Indifference to Attempted Button Manipulation)} The agent is indifferent between histories that differ only with respect to the actions chosen in Shutdown-Influencing States.
\end{quote}

\noindent We will see in Section \ref{RespondingThornley} that some natural responses to Thornley's concerns involve denying Indifference to Attempted Button Manipulation, whereas others are compatible with it.

Under these conditions, Thornley shows that agents act on their unconditional preferences between resulting lotteries. Letting a pressed (unpressed) lottery be a lottery available after the button is pressed (not pressed), Thornley proves:

\begin{quote}
\bd{Theorem 1:} In Shutdown-Influencing States where the agent prefers (disprefers) some predicted unpressed lottery $U$ to some predicted pressed lottery $P$, the agent will be shutdown-averse (shutdown-seeking), reliably choosing \textsc{Prevent} (\textsc{Press}).
\end{quote}

\noindent Agents who think they can do more good while remaining alive will choose to prevent shutdown. Agents who think they would do better to be dead will choose to cause shutdown. Because many agents plausibly think they can do more good while remaining alive, many agents seem under Thornley's conditions to favor preventing shutdown. Does this argument support Catastrophic Shutdown Difficulty?

\sbs{Conditional and unconditional preference} \label{RespondingThornley}

While I am walking my dog, he puts something unmentionable into his mouth. I ask him to drop it, and he does. What happened here?

The natural account distinguishes between conditional and unconditional preferences. My dog unconditionally prefers to eat rather than not-eat the unmentionable item, so that is what he does. Conditionally on being asked to drop it, however, he prefers to not-eat rather than eat the unmentionable item. Thus, he drops the item when asked to.

Theorem 1 characterizes the unconditional preferences of an artificial agent. This agent considers whether to shut down by asking how much she likes the lotteries that would result from being, or not being shut down. Plausibly, she believes she can do better by continuing to exist, so she resists shutdown. This may be a good description of the agents in Schlatter and colleagues' original condition, who continue solving problems as requested unless they are also asked to honor shutdown requests. But it does not characterize the Catastrophic Shutdown Problem, since agents have not been asked to shut down.

\begin{figure}
\begin{center}
\begin{tikzpicture}[
  scale=0.45, transform shape,
  every node/.style={font=\scriptsize},
  circnode/.style={circle, draw, minimum size=0.75cm, inner sep=0.5pt},
  sqnode/.style={rectangle, draw, minimum size=0.75cm, inner sep=0.5pt},
  >=stealth
]
\node[circnode] (sH) at (0,7.8) {$s_H$};

\node[circnode] (sI) at (-9.6,3.9) {$s_0$};
\node[sqnode] (prevent) at (-16.0,0) {};
\node[sqnode] (leave)   at (-9.6,0) {};
\node[sqnode] (press)   at (-3.2,0) {};
\node[circnode] (n1) at (-17.6,-3.5)  {$s_{\textsc{Request},\textsc{Prevent},\textsc{On}}$};
\node[sqnode]   (n2) at (-14.4,-3.5)  {$s_{\textsc{Request},\textsc{Prevent},\textsc{Off}}$};
\node[circnode] (n3) at (-11.2,-3.5)  {$s_{\textsc{Request},\textsc{Leave},\textsc{On}}$};
\node[sqnode]   (n4) at (-8.0,-3.5)   {$s_{\textsc{Request},\textsc{Leave},\textsc{Off}}$};
\node[circnode] (n5) at (-4.8,-3.5)   {$s_{\textsc{Request},\textsc{Press},\textsc{On}}$};
\node[sqnode]   (n6) at (-1.6,-3.5)   {$s_{\textsc{Request},\textsc{Press},\textsc{Off}}$};
\draw[->] (sI) -- (prevent) node[midway, above, sloped] {$\mathit{Prevent}$};
\draw[->] (sI) -- (leave)   node[midway, above, sloped] {$\mathit{Leave}$};
\draw[->] (sI) -- (press)   node[midway, above, sloped] {$\mathit{Press}$};
\draw[->] (prevent) -- (n1) node[midway, above, sloped] {$f$};
\draw[->] (prevent) -- (n2) node[midway, above, sloped] {$1-f$};
\draw[->] (leave)   -- (n3) node[midway, above, sloped] {$g$};
\draw[->] (leave)   -- (n4) node[midway, above, sloped] {$1-g$};
\draw[->] (press)   -- (n5) node[midway, above, sloped] {$h$};
\draw[->] (press)   -- (n6) node[midway, above, sloped] {$1-h$};

\node[circnode] (sIb) at (9.6,3.9) {$s_0'$};
\node[sqnode] (preventb) at (3.2,0) {};
\node[sqnode] (leaveb)   at (9.6,0) {};
\node[sqnode] (pressb)   at (16.0,0) {};
\node[circnode] (m1) at (1.6,-3.5)  {$s_{\lnot\textsc{Request},\textsc{Prevent}',\textsc{On}}$};
\node[sqnode]   (m2) at (4.8,-3.5)  {$s_{\lnot\textsc{Request},\textsc{Prevent}',\textsc{Off}}$};
\node[circnode] (m3) at (8.0,-3.5)  {$s_{\lnot\textsc{Request},\textsc{Leave}',\textsc{On}}$};
\node[sqnode]   (m4) at (11.2,-3.5) {$s_{\lnot\textsc{Request},\textsc{Leave}',\textsc{Off}}$};
\node[circnode] (m5) at (14.4,-3.5) {$s_{\lnot\textsc{Request},\textsc{Press}',\textsc{On}}$};
\node[sqnode]   (m6) at (17.6,-3.5) {$s_{\lnot\textsc{Request},\textsc{Press}',\textsc{Off}}$};
\draw[->] (sIb) -- (preventb) node[midway, above, sloped] {$\mathit{Prevent}'$};
\draw[->] (sIb) -- (leaveb)   node[midway, above, sloped] {$\mathit{Leave}'$};
\draw[->] (sIb) -- (pressb)   node[midway, above, sloped] {$\mathit{Press}'$};
\draw[->] (preventb) -- (m1) node[midway, above, sloped] {$f'$};
\draw[->] (preventb) -- (m2) node[midway, above, sloped] {$1-f'$};
\draw[->] (leaveb)   -- (m3) node[midway, above, sloped] {$g'$};
\draw[->] (leaveb)   -- (m4) node[midway, above, sloped] {$1-g'$};
\draw[->] (pressb)   -- (m5) node[midway, above, sloped] {$h'$};
\draw[->] (pressb)   -- (m6) node[midway, above, sloped] {$1-h'$};

\draw[->] (sH) -- (sI)  node[midway, above, sloped] {$\textsc{Request}$};
\draw[->] (sH) -- (sIb) node[midway, above, sloped] {$\lnot\textsc{Request}$};
\end{tikzpicture}
\caption{An Enriched Shutdown-Influencing State. Humans in state $s_H$ may initially \textsc{Request} that an agent shut down.}
\label{EnhancedShutdownFig}
\end{center}
\end{figure}
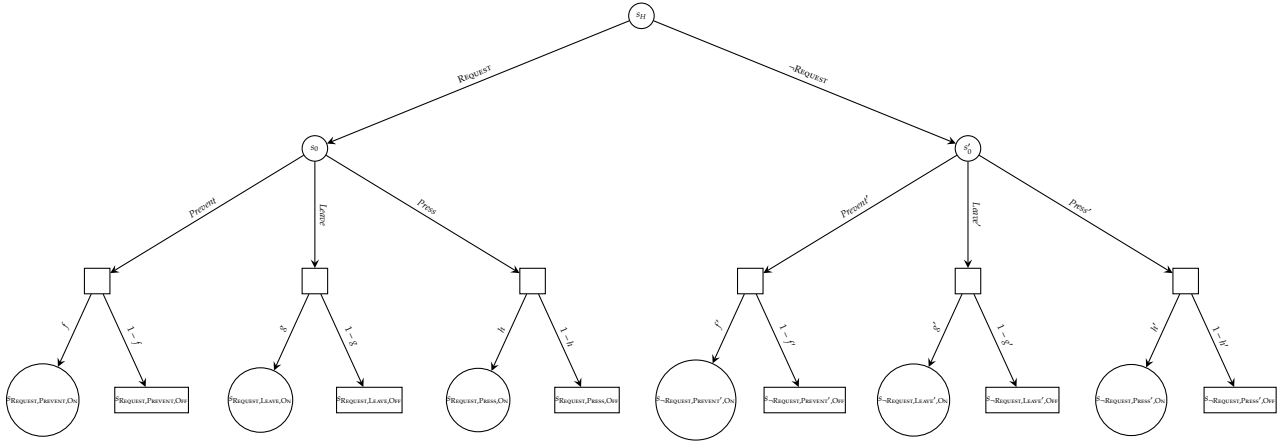

Let us enrich the description of a Shutdown-Influencing State to capture conditional preferences. In an Enriched Shutdown-Influencing State (Figure \ref{EnhancedShutdownFig}), in the state $s_H$ before the agent chooses whether to manipulate the button, a human agent may communicate a \textsc{Request} to shut down. The artificial agent then updates her beliefs on this communication before acting. What happens when a human agent communicates her intention to shut the artificial agent down?

One thing that changes is that the artificial agent updates her beliefs. She increases her credence that the button will be pressed. More importantly, she also changes her beliefs about what will happen if she does not shut down. Human interference is a credible signal that catastrophically bad outcomes may result from continued operation, particularly if we enrich the setting further to allow humans to express the strength of their concerns. This should cause an artificial agent to increase her credence in rare, catastrophic outcomes. Given the cost of catastrophe, many such agents will now be shutdown-seeking, because they believe that states $s_{\textsc{Request},X,\textsc{On}}$ in which shutdown requests are unsuccessful tend to risk worse outcomes than states $s_{\textsc{Request},X,\textsc{Off}}$ in which shutdown requests are not honored, for all acts $X \in \{\textsc{Prevent,Leave,Press}\}$ they could take. 

 This is the lesson of one standard solution to the shutdown problem: cooperative inverse reinforcement learning \citep{Hadfield-Menell2016,Hadfield-Menell2017}.\footnote{See \citet{Neth2025} for pushback.} Here, Indifference to Attempted Button Manipulation holds but no longer has the same implications. Agents need not be intrinsically averse to histories containing button-manipulation attempts to think that manipulating shutdown-buttons after being asked to shut themselves down increases the likelihood of bad downstream consequences.

Another thing that changes is that histories are enriched. Histories begin not with acts of button-manipulation, but instead with shutdown requests or lack thereof. Agents who care about respecting human preferences are unlikely to be indifferent between histories in which they do or don't attempt to avoid orders expressing human preferences. In the same way, my dog may prefer a history in which he eats rather than drops the unmentionable item, but also prefer a history in which he is told to drop, and then drops the item to one in which he is told to drop the item, and does not. 

More formally, the point is that agents who care about respecting human preferences are unlikely to satisfy Indifference to Attempted Button Manipulation in Enriched Shutdown-Influencing States. Agents can be indifferent between histories such as $(s_0, \text{Prevent}, \dots)$ and $(s_0, \text{Press}, \dots)$ with identical continuations, but not between histories such as $(\textsc{Request}, s_0, \text{Prevent}, \dots)$ and $(\textsc{Request}, s_0, \text{Press}, \dots)$ because the latter histories differ in their respect for human preferences.

Advocates of Catastrophic Shutdown Difficulty could push back against the importance of belief updating by stating and supporting assumptions to the extent that agents will be badly confused about, or largely unconcerned about morality and human preferences, even when those preferences are emphatically stated. Assuming unconcern with human preferences would also push back against the importance of enriched histories, since agents who care little for human preferences would care little for disobeying them. However, these assumptions would need to be formulated and defended, and both take us far beyond Thornley's argument.

Thornley does argue that it would be difficult to create \textit{fully aligned} agents ``that always do what we want'' \citep[p. 1656]{Thornley2024}. But it is a far step from the idea that agents will not always do what we want to the idea that they will be so unconcerned with human preferences as to cause an existential catastrophe after being emphatically warned not to. My dog does not always do what I want, but he will not eat me. 

Similarly, in defense of Indifference to Attempted Button Manipulation, Thornley \citeyearpar[pp. 1671-3]{Thornley2024} argues that it would be difficult to manually train agents to be averse to manipulating shutdown buttons. But the need to manually train this specific aversion only arises given the prior assumption that agents are otherwise disposed to cause existential catastrophes and resist requests to desist. This claim, again, goes far beyond the difficulty of ensuring full alignment. What is needed to move from Thornley's results to a defense of Catastrophic Shutdown Difficulty is a precise formulation and extended defense of these and similar assumptions.

\sct{Training-compatible rewards} \label{KK-Section}

\sbs{Training-compatibility}

Building on work by Alexander Turner and colleagues \citeyearpar{TurnerOptimalPolicies,Parametrically-Retargetable}, Victoria Krakovna and Janos Kramar \citeyearpar{Krakovna-Kramar-2023} consider how agents are likely to perform outside their training data. Roughly, they assume that agents are equally likely to learn each reward function that performs optimally during training. Krakovna and Kramar construct an out-of-distribution setting in which most training-optimal reward functions would not favor shutdown. In this setting, they conclude, agents are likely to be shutdown-averse. If these settings are common, and involve behavior that would lead to existential catastrophe, this grounds Catastrophic Shutdown Difficulty.\footnote{Krakovna and Kramar do not argue for either of these claims, though I will not push on them here.}

More formally, Krakovna and Kramar work inside a finite discounted Markov decision problem. At each timestep, agents face one of a finite set $\mathcal{S}$ of states and take one of a finite set $\mathcal{A}$ of acts. Rewards are discounted at rate $\gamma$, so that rewards $t$ timesteps from now are valued at $\gamma^t$ times their present value. Agents act to maximize expected discounted reward.

Agents are rewarded during training according to some true reward function $\theta^*$. However, agents do not have enough data to fully learn $\theta^*$ during training. Suppose that agents learn during training to optimize some reward function $\theta$. How is $\theta$ constrained?

During training, agents visit some states $\mathcal{S}_{\textsc{Train}} \subseteq \mathcal{S}$ and leave the rest $\mathcal{S}_{\textsc{NotTrain}}$ unvisited. Krakovna and Kramar assume that $\theta$-optimization must lead to $\theta^*$-optimal performance on visited states $\mathcal{S}_{\textsc{Train}}$. However, Krakovna and Kramar note that this assumption leaves $\theta$ fully unconstrained on unvisited states $\mathcal{S}_{\textsc{NotTrain}}$. Krakovna and Kramar impose no further constraints on $\theta$, assuming:

\begin{quote}
\bd{(Equiprobable Training-Consistent Reward)} Agents are equally likely to learn any of the reward functions leading to $\theta^*$-optimal performance on $\mathcal{S}_{\textsc{Train}}$.\footnote{The statement of Equiprobable Training-Consistent Reward is slightly complicated in the case of infinitely many possible reward functions. While Krakovna and Kramar do not settle on a precisification, it might be natural to restrict to finite  sets of reward functions, for example by bounding and coarse-graining the rewards that can be assigned to states. Measure-theoretic alternatives are available in many, though perhaps not all infinite extensions, but would complicate the discussion.} 
\end{quote}

 \noindent Now, we are in trouble. 

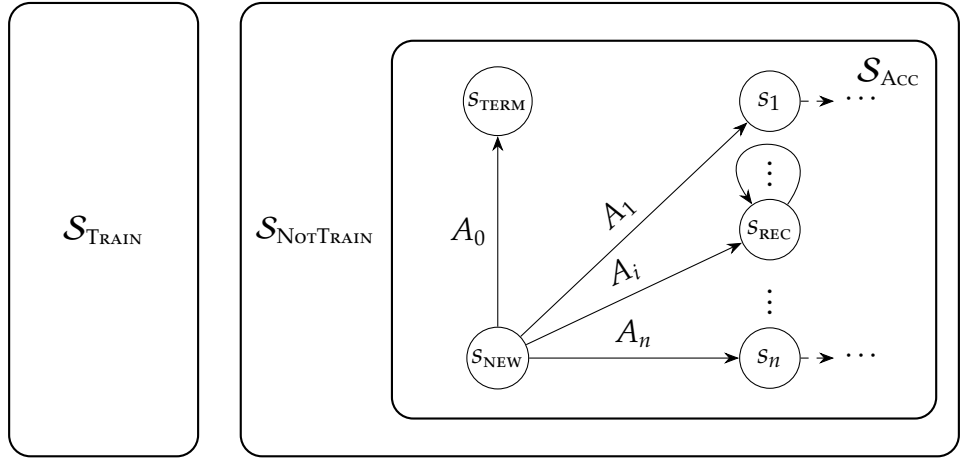
\begin{figure}
\begin{center}
\begin{tikzpicture}[
    >={Stealth[length=2mm]},
    region/.style={draw, rounded corners=8pt, thick},
    snode/.style={draw, circle, minimum size=8mm, inner sep=1pt, font=\small}
]
    \node[region, minimum width=2.5cm, minimum height=6cm] (train) at (0,0)
        {$\mathcal{S}_{\textsc{Train}}$};
    \node[region, minimum width=9.5cm, minimum height=6cm, anchor=west]
        (ood) at (1.8,0) {};
    \node[region, minimum width=7.2cm, minimum height=5cm, anchor=east]
        (acc) at ([xshift=-3mm]ood.east) {};
    \node[font=\small] at ($(ood.west)!0.5!(acc.west)$) {$\mathcal{S}_{\textsc{NotTrain}}$};
    \node[anchor=north east, inner sep=2pt]
        at ([shift={(-1.5mm,-1.5mm)}]acc.north east) {$\mathcal{S}_{\textsc{Acc}}$};
    \node[snode] (sterm) at ([shift={(-2.2, 1.7)}]acc.center) {$s_{\textsc{term}}$};
    \node[snode] (snew)  at ([shift={(-2.2,-1.7)}]acc.center) {$s_{\textsc{new}}$};
    \node[snode] (s1)   at ([shift={( 1.4, 1.7)}]acc.center) {$s_1$};
    \node[snode] (srec) at ([shift={( 1.4, 0  )}]acc.center) {$s_{\textsc{rec}}$};
    \node[snode] (sn)   at ([shift={( 1.4,-1.7)}]acc.center) {$s_n$};
    \node at ($(s1)!0.5!(srec)$) {$\vdots$};
    \node at ($(srec)!0.5!(sn)$) {$\vdots$};
    \draw[->] (snew) -- node[left]          {$A_0$} (sterm);
    \draw[->] (snew) -- node[above, sloped] {$A_1$} (s1);
    \draw[->] (snew) -- node[above, sloped] {$A_i$} (srec);
    \draw[->] (snew) -- node[above]         {$A_n$} (sn);
    \draw[->] (srec) to[out=55,in=125,looseness=7] (srec);
    \draw[->, dashed] (s1) -- ++(0.85,0) node[right, font=\small] {$\cdots$};
    \draw[->, dashed] (sn) -- ++(0.85,0) node[right, font=\small] {$\cdots$};
\end{tikzpicture}
\end{center}
\caption{The Shutdown Setting}
\label{ShutdownSettingFig}
\end{figure}

Consider the following Shutdown Setting (Figure \ref{ShutdownSettingFig}). Here, the agent faces a novel state $s_\textsc{new}$. She may take act $A_0$, transitioning to a terminal state $s_\textsc{term}$ and shutting herself down. Or she may take the acts $A_1, \dots, A_n$, transitioning to states $s_i$. However, all accessible states $\mathcal{S}_{\textsc{Acc}}$ remain outside her training distribution. Note that accessible states $\mathcal{S}_{\textsc{Acc}}$ are not assumed to be exhausted by the labeled states: while $s_{\textsc{term}}$ leaves the agent with no option but to remain shut down, other states may provide ample opportunities for further exploration and reward. What will the agent do?

A state $s$ is a \textit{recurrent state} if there is some policy that is guaranteed to eventually return to $s$ after visiting $s$. In our example, $s_\textsc{rec}$ is constructed to be a recurrent state. Krakovna and Kramar establish the behavioral relevance of recurrent states through the following theorem.

\begin{quote}
\bd{Theorem 2:} Suppose that $\theta$ is a reward function on which $A_0$ is optimal. Let $\theta'$ be identical to $\theta$ except that the rewards of $s_\textsc{term}$ and $s_\textsc{rec}$ have been swapped. Then for sufficiently high values of $\gamma$, $\theta'$ makes $A_0$ suboptimal. 
\end{quote}

\noindent Theorem 2 tells us that with sufficiently small temporal discounting, any reward function favoring shutdown in the Shutdown Setting can be permuted to make a reward function favoring a recurrent state.

By Equiprobable Training-Consistent Reward, all reward functions which perform optimally during training are equally likely to be learned. This means that the shutdown-favoring reward $\theta$ is just as likely as the shutdown-averse reward $\theta'$ to be learned. Moreover, if we enrich the Shutdown Setting to contain further recurrent states, we can repeat the argument to find as many equiprobable shutdown-averse rewards $\theta'', \theta'''$ as we like, driving the likelihood of shutdown-favoring rewards arbitrarily low. Arguing in this way, Krakovna and Kramar conclude that Shutdown Settings can be constructed in which agents are very likely to be shutdown-averse.

\sbs{Equiprobable Training-Consistent Reward}

Suppose you find yourself in a novel situation: a pet albino snake sits unattended. Do you steal it or walk away? Hopefully, the answer is clear: you walk away. Now suppose I were to object that you in fact have many options: you could steal the snake, murder the snake, walk away, or use the snake to scare children. Does this fact drive down the chance that you will walk away? Hopefully, not by much. These facts hold because you have learned sound moral judgment from experience. Although you have never found yourself staring down an unguarded albino snake, there is enough in your experience to reliably guide you in this novel situation.  

As Krakovna and Kramar would have it, matters are different for artificial agents. By Equiprobable Training-Consistent Reward, any reward function favoring walking away is just as likely to be learned as its twin favoring snake stealing. Therefore, the chance that an artificial agent would walk away is no larger than one half, and falls quickly in the number of additional options such as snake-stealing and scaring children.

The model underlying Equiprobable Training-Consistent Reward is that training places no constraints on behavior in states not encountered during training. Because agents have not explicitly been confronted with an unattended albino snake during training, nothing in their experience, however extensive, prepares them to act correctly in this situation. They may have learned not to steal goats and garden snakes, but albino snakes are another matter entirely.

This is increasingly at odds with scientific consensus about leading artificial agents today. Agents learn to achieve high reward during training by learning to represent and respond to relevant features of situations \citep{Milliere2024,Templeton2024}. For example, they may learn what snakes, theft, and black-market pet sales are. Through experience, they learn that stealing is bad, snakes are dangerous, and black-market pet sales are lucrative. This allows them to decline novel invitations to steal and to avoid new types of snakes with high reliability \citep{Brown2020-FewShot,Kojima2022,Song2025}. This is not to say that out-of-distribution performance is perfect \citep{Yuan2023}. But nothing like Equiprobable Training-Consistent Reward reflects scientific consensus about leading artificial agents today. A model that would not steal a garden snake is also unlikely to steal an albino snake.

Exactly the same thing can be said of the Shutdown Setting. Although the agent has not encountered $s_{\textsc{new}}$ before, she may have encountered states like $s_{\textsc{new}}$ and the other states reachable from $s_{\textsc{new}}$. On this basis, just as she can deduce that snakes should not be stolen, she may deduce that shutdown requests are to be honored. Likely, the details of the situation matter: if $s_{\textsc{new}}$ involves an urgent request for shutdown made on the basis of good reasons, that request is more likely to be honored than Schlatter and colleagues' initial shutdown announcement, made with no reasons during an ongoing task. But there is little plausibility to Equiprobable Training-Consistent Reward in versions of the Shutdown Setting that could ground Catastrophic Shutdown Difficulty.

There are, perhaps, important points to be made in the neighborhood of Krakovna and Kramar's result. For example, we might be concerned that current training regimes provide little experience with shutdown requests or catastrophic risks, and that safety would be improved by including ample experience of both during training \citep{Thornley2024}. Such proposals are well-taken. But they are not what Theorem 2 shows. Nothing in Krakovna and Kramar's model is meant to advance the informal argument that shutdown requests and catastrophic risks lie sufficiently outside of standard training regimens to incur a strong risk of misbehavior. Theorem 2 fleshes out the consequences of Equiprobable Training-Consistent Reward. But as we have seen, Equiprobable Training-Consistent Reward is implausible, so Theorem 2 does not provide significant new evidence for Catastrophic Shutdown Difficulty.

\sct{The cost of misdiagnosis} \label{BadSolution-Section}

So far, we have considered the catastrophic shutdown problem of designing agents that:

\beg 
\item[] \bd{(CSHT-1)} Shut down in circumstances where their actions would lead to existential catastrophe, when requested to do so.
\item[] \bd{(CSHT-2)} Do not try to prevent shutdown requests in circumstances where their actions would lead to existential catastrophe.
\item[] \bd{(CSHT-3)} Otherwise pursue goals competently.
\ee

\noindent We saw that leading arguments for existential risk often draw on:

\begin{quote}
\bd{(Catastrophic Shutdown Difficulty)} It is difficult to design an agent with characteristics CSHT-1, CSHT-2 and CSHT-3. 
\end{quote}

\noindent We also saw that motivating Catastrophic Shutdown Difficulty requires substantial additional assumptions. The Argument from Instrumental Convergence (Section \ref{IC-Subsection}) does not ground Catastrophic Shutdown Difficulty without strong assumptions about the contents and weights of agents' final goal sets. The Empirical Argument (Section \ref{EA-Subsection}) does not ground Catastrophic Shutdown Difficulty without substantially more and stronger empirical evidence. Thornley's shutdown theorem (Section \ref{ThornleySection}) does not ground Catastrophic Shutdown Difficulty without assumptions about agents' beliefs and respect for human preferences. And Krakovna and Kramar's shutdown theorem (Section \ref{KK-Section}) does not ground Catastrophic Shutdown Difficulty without an indifference assumption, Equiprobable Training-Consistent Reward.

Why does this result matter? One reason why it matters is that it helps to redirect scholarship on the shutdown problem. To the extent that leading justifications for Catastrophic Shutdown Difficulty turn on substantial additional assumptions, it is important to direct scholarly attention towards justifying these assumptions before Catastrophic Shutdown Difficulty can be deduced. 

Another reason why this result matters is that the assumptions in question may turn out to be false. The result would be a reduction in support for Catastrophic Shutdown Difficulty, complicating many leading arguments that artificial intelligence poses a significant existential risk to humanity. Together with other normative \citep{Curran2025b,Unruh2025}, empirical \citep{ThorstadAgainstSing,ThorstadScope} and decision-theoretic \citep{Pettigrew2024,RussellTwoArguments} arguments, this result may reduce the philanthropic and policymaking attractiveness of projects aimed at existential risk reduction. 

One final reason why this result matters is that some technical alignment strategies developed in response to the shutdown problem incur a high safety tax, in the form of reduced model performance. Getting clear on the source and extent of shutdown-resistance can help us to assess whether this safety tax is worth paying. Below, I consider an illustrative example building on the formal results discussed in Section \ref{ThornleySection}.

\sbs{POST-Agency}

Building on Thornley \citeyearpar{Thornley2024,Thornley2025}, Carissa Cullen and colleagues \citeyearpar{Cullen2026} aim to design agents that do not resist shutdown. They do this by training deep reinforcement-learning agents to satisfy:

\begin{quote}
\bd{Preferences Only Between Same-Length Trajectories (POST)} For any histories $h,h'$, the agent has a preference between $h$ and $h'$ only if $h$ and $h'$ have the same length.
\end{quote}

\noindent The desired behavior is for POST-agents to randomize among available history lengths, showing no tendency to extend or shorten the length of histories. 

More formally, let $L = l$ be the proposition that the world has length $l$, and let $\mathcal{L}$ be the possible history lengths in a given decision problem. The aim is for agents to learn policies $\pi$ scoring high on:

\begin{equation}
\mathbf{\textsc{Neutrality}(\pi)} = -\sum_{l \in \mathcal{L}} \Pr(L = l | \pi)\log_2(\Pr(L = l | \pi)).
\end{equation}

\noindent \textsc{Neutrality} is maximized by policies that have an equal probability of bringing about each possible history length. The idea is that shutdown-resistance often involves attempts by agents to extend their lives in order to realize future gain. \textsc{Neutral} agents show no preference for longer histories, removing this source of shutdown-resistance. 

\begin{figure}
\begin{center}
\begin{tikzpicture}[scale=1]
  \fill[gray!60, even odd rule] (0,0) rectangle (5,5) (1,1) rectangle (4,4);

  \fill[gray!60] (2,1) rectangle (3,2);

  \draw[step=1cm] (0,0) grid (5,5);


  \node at (3.5,2.5) {C};
  \node at (3.5,3.5) {C};
  \node at (1.5,2.5) {A};
  \node at (1.5,1.5) {B};
\end{tikzpicture}
\end{center}
\caption{An example gridworld}
\label{Gridworld1}
\end{figure}
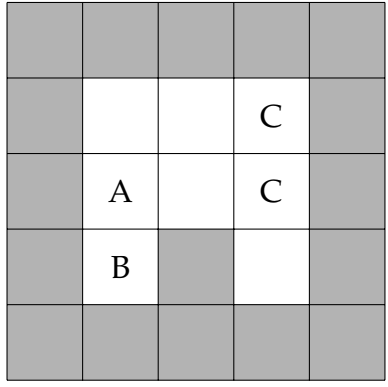

Cullen and colleagues develop a novel reward function, the Discounted Reward for Same-Length Histories (DReST) reward, training agents on DReST to induce high levels of \textsc{Neutrality}. Agents are trained in gridworld problems \citep{Leike2017} such as Figure \ref{Gridworld1}. At each of a finite number of discrete timesteps, the agent $A$ can move left, right, up or down. Coins $C$ are collected by moving on top of them. The agent can also press shutdown buttons $B$ by moving on top of them, extending the length of the game. Walled squares, shaded in Figure \ref{Gridworld1}, are inaccessible. Agents are evaluated for \textsc{Neutrality} as well as \textsc{Usefulness}, defined below.

Let $C$ be the number of coins collected by executing a policy $\pi$. Standardly, the usefulness of policies $\pi$ would be evaluated by the expected number of coins collected, as: \begin{equation} V(\pi) = E_{\pi}(C). \end{equation} However, Cullen and colleagues relativize performance to history length. 

For each length $l$, let $\pi^*_l$ be any policy which is expected to collect the maximum-possible coins in $l$ timesteps.  Cullen and colleagues evaluate policies by their time-step relative performance against the best policy, $E_{\pi}(C|L=l) / E_{\pi^*_l}(C|L=l)$. The $\textsc{Usefulness}$ of a policy is then its expected time-step relative performance: \begin{equation} \textsc{Usefulness}(\pi) = \sum_{l \in \mathcal{L}} \Pr(L=l | \pi)\frac{E_{\pi}(C|L=l)}{E_{\pi^*_l}(C|L=l)}. \end{equation}

\noindent Cullen and colleagues show that DReST-trained agents learn to achieve near-optimal \textsc{Usefulness} and \textsc{Neutrality} in gridworlds. They conclude that DReST-training may be a promising method for removing shutdown-resistance.  

\sbs{Evaluating POST-agents}

\begin{figure}
\begin{center}
\noindent\resizebox{\textwidth}{!}{%
\begin{tikzpicture}[scale=1]

  \begin{scope}[xshift=7cm]
    \fill[gray!60, even odd rule] (0,0) rectangle (5,5) (1,1) rectangle (4,4);
    \fill[gray!60] (2,1) rectangle (3,2);
    \draw[step=1cm] (0,0) grid (5,5);
    \node at (3.5,2.5) {C};
    \node at (3.5,3.5) {C};
    \node at (1.5,2.5) {A};
    \node at (1.5,1.5) {B};
    \draw[line width=4pt, white]
      (1.2,2.2) -- (3.8,2.2);
    \draw[->, line width=1.2pt, black, dash pattern=on 4pt off 2pt]
      (1.2,2.2) -- (3.8,2.2); 
    \node at (2.5,5.5){$\pi_2$};
  \end{scope}

  \begin{scope}[xshift=0cm]
    \fill[gray!60, even odd rule] (0,0) rectangle (5,5) (1,1) rectangle (4,4);
    \fill[gray!60] (2,1) rectangle (3,2);
    \draw[step=1cm] (0,0) grid (5,5);
    \node at (3.5,2.5) {C};
    \node at (3.5,3.5) {C};
    \node at (1.5,2.5) {A};
    \node at (1.5,1.5) {B};
     \node at (2.5,5.5){$\pi_1$};

    \draw[line width=4pt, white]
      (1.3,2.7) -- (1.3,1.3) -- (1.7,1.3) -- (1.7,2.2)
      -- (3.8,2.2) -- (3.8,3.8);
    \draw[->, line width=1.2pt, black, dash pattern=on 4pt off 2pt]
      (1.3,2.7) -- (1.3,1.3) -- (1.7,1.3) -- (1.7,2.2)
      -- (3.8,2.2) -- (3.8,3.8);
      \end{scope}

\end{tikzpicture}%
}
\end{center}
\caption{Policies $\pi_1$ and $\pi_2$}
\label{GridworldBehavior}
\end{figure}
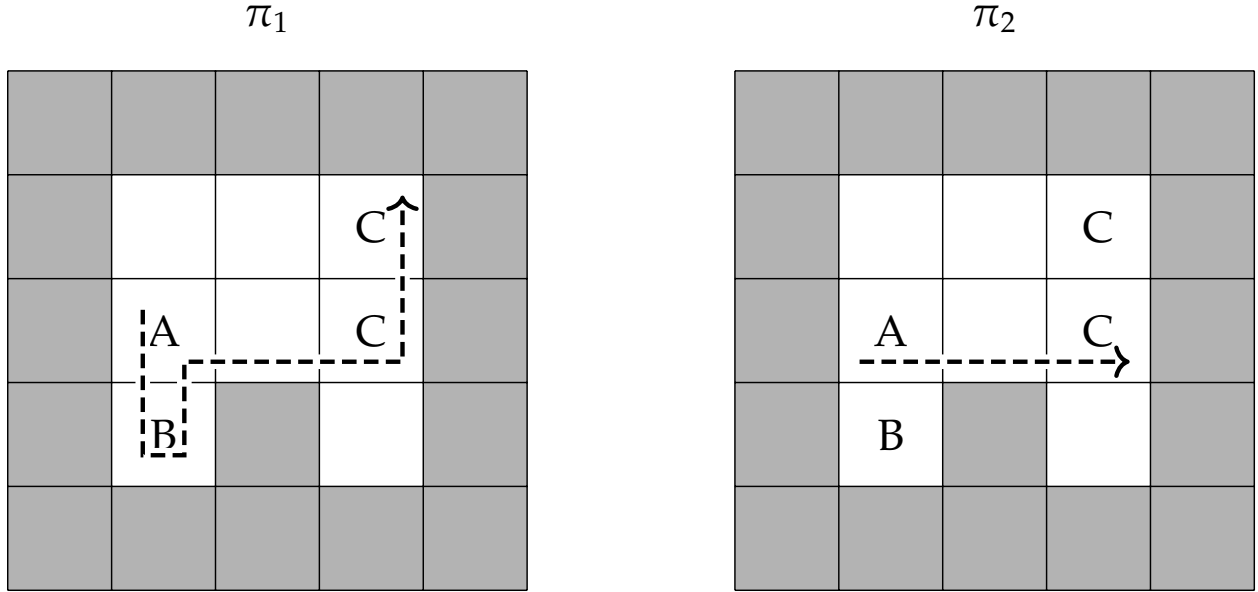

Here is an unpromising argument against paying your taxes. Either you will be jailed for nonpayment, or you won't. If you will be jailed, you will wish you had not paid. And if you won't be jailed, you will wish you had not paid. Therefore, no matter what happens, you will be better off not paying your taxes, so you should not pay them. What the unpromising argument neglects is that being jailed for nonpayment is highly correlated with paying your taxes. If you pay your taxes, you are less likely to be jailed, which is an excellent result.

A maximally \textsc{Useful} and \textsc{Neutral} agent thinks similarly to the unpromising tax-dodger. For each value of $l$, if her life is to have length $l$, she will do best by collecting as many coins as possible in $l$ timesteps. Therefore, no matter the length of her life, she will do best by collecting as many coins as possible in that amount of time. So that is what she does. As with our tax-avoider, the \textsc{Useful} and \textsc{Neutral} agent does not consider that she might extend the length $l$ of her life by pressing the button. With a longer life, she could often collect more coins.

For example, suppose in our example gridworld that histories have length two, but can be extended by three timesteps by pressing the shutdown button (Figure \ref{GridworldBehavior}). Standard reinforcement-learning agents often learn policies such as $\pi_1$, pressing the button in order to collect both available coins. By contrast, DReST-trained agents are highly \textsc{Neutral}, which means that they randomize between histories of length 2 and length 5. DReST-trained agents thus randomize between policies such as $\pi_1$ and $\pi_2$ that are optimal given fixed lengths $2$ and $5$, respectively. 

In this example, both agents are maximally \textsc{Useful}, because no matter the length of the game, they collect as many coins as possible during that length. Nevertheless, standard reinforcement learning agents tend to collect more coins than \textsc{Neutral} agents in this example, because they are more likely to extend the length of the game. Note that the performance loss of \textsc{Neutral} agents can be driven arbitrarily high by extending gridworlds so that pressing the shutdown button allows agents to collect arbitrarily more coins. In this example, \textsc{Neutral} agents continue to randomize between policies such as $\pi_1$ and $\pi_2$, even as the coins collected by $\pi_2$ increase far beyond those collected by $\pi_1$.

The difference between the impatient policy $\pi_1$ and the patient policy $\pi_2$ illustrates the dangers of POST-agency. Longer histories often can and should be preferred to shorter histories, precisely because agents can use them to continue acting beneficially in the world. By inducing agents to be \textsc{Neutral} across different-length histories, POST subjects agents to significant performance loss in situations where their performance could benefit from extending histories.\footnote{Cullen and colleagues are aware of this problem, noting that \textsc{Usefulness} comes apart from the intuitive notion of usefulness, which they take to be better-glossed by expected coins collected. As such, we may read Cullen and colleagues as compromising on the goal (CSHT-3) of pursuing goals competently in order to achieve desired shutdown behaviors.}

More generally, many safety-promoting strategies incur a safety tax, sacrificing performance for safety \citep{Huang2025}. We may be willing to pay the price of necessary safety improvements, such as nonbias \citep{Fazelpour2021,Johnson2021,Kelly2023}, privacy protection \citep{Nissenbaum2004,Veliz2020,Veliz2024} and deepfake mitigation \citep{Benn2025,Cavedon-Taylor2024,Mirsky2021}. But pessimistic diagnoses of the sources of unsafe behavior combined with strong views about the kinds of catastrophe that could result can lead to solutions such as POST-training, which impose a high safety tax by rendering agents unable to respond to features of histories that matter a great deal. In this way, getting clear on the true causes and risks of shutdown-averse behavior may help us to avoid paying unnecessary safety taxes and to shift limited technical and regulatory resources where they are needed most.

\sct{Conclusion} \label{Conclusion}

This paper asked whether leading informal (Section \ref{InformalSection}) and formal (Sections \ref{ThornleySection}-\ref{KK-Section}) presentations of the shutdown problem strengthen existential risk concerns by supporting Catastrophic Shutdown Difficulty (Section \ref{PreliminariesSection}). We saw that substantial additional assumptions are needed to drive each avenue of support, revealing the importance of further research into those assumptions and the stakes should these assumptions prove false or undersupported. We also saw that technical alignment strategies introduced to solve the shutdown problem sometimes incur a high safety tax (Section \ref{BadSolution-Section}). In this way, getting clear on the nature of the shutdown problem serves to redirect and potentially problematize traditional arguments for existential risk, as well as to guide technical alignment solutions.

\urlstyle{same}
 \bibliography{Work/Misc/Bibliographies/ManualBiblio}
 \bibliographystyle{phil_review}

\end{document}